\pdfoutput=1

\documentclass[11pt]{article}

\usepackage{acl}

\usepackage{times}
\usepackage{latexsym}
\usepackage{bm}
\usepackage{amsthm,amsmath,amssymb}
\usepackage{mathrsfs}
\usepackage{epsfig}
\usepackage{graphicx}
\usepackage{booktabs}
\usepackage{verbatim}
\usepackage{enumitem,kantlipsum}
\usepackage{tipa}
\usepackage{algorithm}
\usepackage{algorithmicx}
\usepackage{algpseudocode}

\usepackage{cleveref}

\usepackage[T1]{fontenc}

\usepackage[utf8]{inputenc}

\usepackage{microtype}

\usepackage{color} 
\definecolor{mypink2}{RGB}{219, 48, 122}
\definecolor{orange}{RGB}{255, 147, 00}
\definecolor{jrcolor}{RGB}{100, 150, 225}
\definecolor{jrcomment}{RGB}{70, 200, 150}

\algnewcommand\algorithmicinput{\textbf{Input:}}
\algnewcommand\INPUT{\item[\algorithmicinput]}
\algnewcommand\algorithmicoutput{\textbf{Output:}}
\algnewcommand\OUTPUT{\item[\algorithmicoutput]}

\crefformat{section}{\S#2#1#3} 
\crefformat{subsection}{\S#2#1#3}
\crefformat{subsubsection}{\S#2#1#3}

\title{A Survey on   Retrieval-Augmented Text Generation}

\author{Huayang Li$^{\heartsuit,}$\thanks{~~All authors contributed equally.}~\quad Yixuan Su$^{\spadesuit,*}$\quad Deng Cai$^{\diamondsuit,*}$\quad Yan Wang$^{\clubsuit,*}$\quad Lemao Liu$^{\clubsuit,*}$\\
$^\heartsuit$Nara Institute of Science and Technology \ \ \ \ \  $^\spadesuit$University of Cambridge\\
$^\diamondsuit$The Chinese University of Hong Kong \ \ \ \ \ $^\clubsuit$Tencent AI Lab
\\ {\tt li.huayang.lh6@is.naist.jp, ys484@cam.ac.uk}\\ {\tt thisisjcykcd@gmail.com, brandenwang@tencent.com} \\ {\tt lemaoliu@gmail.com}\\
}

\begin{document}
\maketitle
\begin{abstract}
Recently, retrieval-augmented text generation attracted increasing attention of the computational linguistics community. Compared with conventional generation models, retrieval-augmented text generation has remarkable advantages and particularly has achieved state-of-the-art performance in many NLP tasks. This paper aims to conduct a survey about retrieval-augmented text generation. It firstly highlights the generic paradigm of retrieval-augmented generation, and then it reviews notable approaches according to different tasks including dialogue response generation, machine translation, and other generation tasks. Finally, it points out some promising directions on top of recent methods to facilitate future research.
\end{abstract}

\section{Introduction}
Retrieval-augmented text generation, as a new text generation paradigm that fuses emerging deep learning technology and traditional retrieval technology, has achieved state-of-the-art (SOTA) performance in many NLP tasks and attracted the attention of the computational linguistics community \cite{weston-etal-2018-retrieve,dinan2018wizard,cai-etal-2021-neural}. Compared with generation-based counterpart, this new paradigm has some remarkable advantages: 1) The knowledge is not necessary to be implicitly stored in model parameters, but is explicitly acquired in a plug-and-play manner, leading to great scalibility; 2) Instead of generating from scratch, the paradigm generating text from some retrieved human-written reference, which potentially alleviates the difficulty of text generation. 


This paper aims to review many representative approaches for retrieval-augmented text generation tasks including dialogue response generation~\cite{weston-etal-2018-retrieve}, machine translation~\cite{gu2018search} and others~\cite{hashimoto2018retrieve}. We firstly present the generic paradigm of retrieval-augmented generation as well as three key components under this paradigm, which are retrieval sources, retrieval metrics 
and generation models.
\begin{figure*}[h]     
  \center{\includegraphics[width=\textwidth] {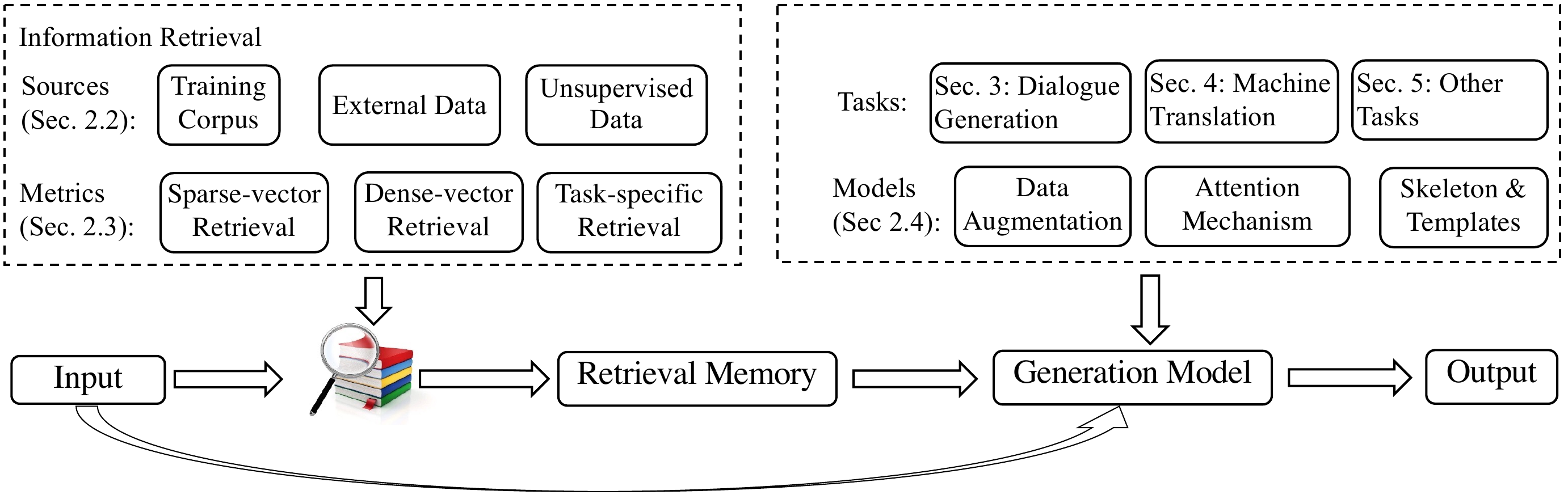}}        
  \caption{
  The overview of this survey.}
  \label{img:overview}
\end{figure*}

Then, we introduce notable methods about retrieval-augmented generation, which are organized with respect to different tasks. Specifically, on the dialogue response generation task, exemplar/template retrieval as an intermediate step has been shown beneficial to informative response generation \cite{weston-etal-2018-retrieve,wu2019response,cai2019skeleton,cai-etal-2019-retrieval}. In addition, there has been growing
interest in knowledge-grounded generation exploring different forms of knowledge such as knowledge bases and external documents \cite{dinan2018wizard,zhou2018dataset,lian2019learning,li2019incremental,qin2019conversing,wu2021controllable,zhang2021joint}. 
On the machine translation task, we summarize the early work on how the retrieved sentences (called translation memory) are used to improve statistical machine translation (SMT)~\cite{koehn-etal-2003-statistical} models~\cite{simard2009phrase,koehn2010convergence} and in particular, we intensively highlight several popular methods to integrating translation memory to NMT models~\cite{gu2018search,zhang2018guiding,xu2020boosting,he2021fast}. 
We also review the applications of retrieval-augmented generation in other generation tasks such as abstractive summarization \cite{peng2019text}, code generation \cite{hashimoto2018retrieve}, paraphrase \cite{kazemnejad-etal-2020-paraphrase,su-etal-2021-keep}, and knowledge-intensive generation \cite{lewis2020retrieval}. 
Finally, we also point out some promising directions on retrieval-augmented generation to push forward the future research. 

\section{Retrieval-Augmented Paradigm}
In this section, we first give a general formulation of retrieval-augmented text generation. Then, we discuss three major components of the retrieval-augmented generation paradigm, including the retrieval source, retrieval metric and integration methods.
\subsection{Formulation}
Most text generation tasks can be formulated as a mapping from input sequence $\boldsymbol{x}$ to output sequence $\boldsymbol{y}: \boldsymbol{y} = f(\boldsymbol{x})$. For instance, $\boldsymbol{x}$ and $\boldsymbol{y}$ could be the dialogue history and the corresponding response for dialogue response generation, the text in the source language and the translation in the target language for machine translation, and so on.


Recently, some researchers suggest to endow models the capability to access external memory via some information retrieval techniques, so that they can acquire more information in the generation process \cite{gu2018search, weston-etal-2018-retrieve, cai-etal-2019-retrieval}. The retrieval-augmented generation can be further formulated as:
\begin{equation}
    \boldsymbol{y} = f(\boldsymbol{x}, \boldsymbol{z})
\end{equation}
where $\boldsymbol{z} = \{\langle \boldsymbol{x}^r, \boldsymbol{y}^r\rangle\}$ is a set of relevant instances retrieved  from the original training set or external datasets. The main idea of this paradigm is that $\boldsymbol{y}^r$ may benefit the response generation, if $\boldsymbol{x}^r$ (or $\boldsymbol{y}^r$) is similar (or relevant) to the input $\boldsymbol{x}$. It is worth noting that $\boldsymbol{x}^r=\emptyset$ when unsupervised retrieval sources are used. In general, the retrieval memory can be retrieved from three kinds of sources: the training corpus, external datasets in the same format with the training corpus, and large-scale unsupervised corpus (\cref{sources}). Metrics that evaluate the relevance between text are varied as well, in \cref{metrics} we divided them into three categories: sparse-vector retrieval, dense-vector retrieval, and training-based  retrieval. Finally, how to integrate the retrieval memory to the generation model is also significant, we also introduce some popular integration approaches in  \cref{integration_fashion}.

\subsection{Retrieval Sources }
\label{sources}
\paragraph{Training Corpus} Most previous studies search the external memory from its \textit{training corpus}~\cite{song2016two, gu2018search, weston-etal-2018-retrieve}. In the inference time, retrieved examples with high relevant scores could be regarded as extra references and reduce model's uncertainty in generation. The main motivation of those works is to to store  knowledge not only in the model parameters but also in an explicit and accessible form, making the model be able to re-access it during inference.

\paragraph{External Data} Some researchers also propose to retrieval relevant samples from \textit{external datasets} \cite{su2021prototype, DBLP:conf/emnlp/XiaoPLWSC21}. In these studies, the retrieval pool is different with the training corpus, which can further provide additional information that are not contained in the training corpus. This is especially beneficial for applications  such as domain adaptation and knowledge update. For example, \citet{khandelwal2020nearest,zheng2021adaptive} employ the in-domain dataset as the external memory to achieve fast domain adaptation for machine translation.

\paragraph{Unsupervised Data} One limitation for previous two sources is that the datasets have to be supervised datasets consisting of aligned input-output pairs. For machine translation, \newcite{cai-etal-2021-neural} propose a cross-lingual retriever to directly retrieve target sentence from \textit{unsupervised corpus} (i.e., monolingual corpus in the target language). The main idea is aligning source-side sentences and the corresponding target-side translations in a dense vector space, i.e., aligning $\boldsymbol{x}$ and $\boldsymbol{y}^r$ when $\boldsymbol{x}^r$ is absent. As a result, the retriever directly connects the dots between the source-side input and target-side translations, enabling monolingual data in the target language to be used alone as memories.

\subsection{Retrieval Metrics}
\label{metrics}
\paragraph{Sparse-vector Retrieval} Given an input sequence $\boldsymbol{x}$ and a retrieval corpus, retrieval model aims to retrieve a set of relevant examples $\boldsymbol{z} = \{\langle \boldsymbol{x}^r, \boldsymbol{y}^r\rangle\}$ from the corpus. When a supervised corpus is used, $\{\langle \boldsymbol{x}^r, \boldsymbol{y}^r\rangle\}$ is retrieved by measuring the similarity between $\boldsymbol{x}$ and $\boldsymbol{x}^r$. For similarity measurement, \textit{sparse-vector retrieval} methods such as TF-IDF and BM25~\cite{robertson2009probabilistic} are widely used. They match keywords efficiently with an inverted index. 

\paragraph{Dense-vector Retrieval} However, these methods prefer examples with similar surfaces, and may fail to retrieve examples that are only semantically relevant. To alleviate above problem, some studies~\cite{cao-xiong-2018-encoding} attempt to retrieve in \textit{dense-vector space} instead of the lexical overlap. Recent work \cite{lee2019latent} makes use of pre-trained language models, which encodes the text to low-dimensional dense vectors via BERT-based encoders. The retrieval score are computed via inner products between vectors.

\paragraph{Task-specific Retrieval} Similarity-based retrieval is based on a simple heuristic. That is, the more $\boldsymbol{x}^r$ resembles with $\boldsymbol{x}$, the more likely $\boldsymbol{x}^r$ and $\boldsymbol{y}^r$ will help the generation. However, the most similar one by universal textual similarity does not necessarily serve the best for downstream models. Ideally, the retrieval metric would be learned from the data in a task-dependent way: we wish to consider a memory only if it can indeed boost the quality of final generation. To this end, \newcite{cai-etal-2021-neural} propose to unify the memory retriever and its downstream generation model into a learnable whole. Such memory retrieval is end-to-end optimized for \textit{task-specific} objectives.

\subsection{Integration}
\label{integration_fashion}
\paragraph{Data Augmentation} There are several ways to integrate the retrieved external memory in generation. One straightforward way is \textit{data augmentation}, which constructs some augmented inputs by concatenating spans from $\{\langle \boldsymbol{x}^r, \boldsymbol{y}^r\rangle\}$ with the original input $\boldsymbol{x}$. By training on the augmented inputs, a generation model implicitly leans how to integrate the retrieved information.
Despite the simplicity, this kind of methods works efficiently in lots of tasks \cite{song2016two, weston-etal-2018-retrieve, bulte2019neural}.


\paragraph{Attention Mechanisms} Another integration method is based on \textit{attention mechanisms} \cite{bahdanau2014neural}. The main idea of this fashion is adopting additional encoders (in various architectures) to encode retrieved target sentences, and integrate them through attention \cite{cao-xiong-2018-encoding, gu2018search, bapna2019non}. Since the attention mechanism is becoming \cite{bahdanau2014neural, vaswani2017attention} a key module in lots of NLP models, integrating retrieved memory through attention becomes a very nature and efficient way.  

\paragraph{Skeleton Extraction} In the previous two methods, the downstream generation model learns how to filter out irrelevant or even harmful information from the retrieved examples implicitly. There also exist some works that try to explicitly extract useful information, i.e., \textit{skeleton extraction}, from the retrieved memory \cite{cai2019skeleton, wu2019response, cai-etal-2019-retrieval}. For example, one skeleton should be a part of a whole utterance with irrelevant content masked, and the generation model only integrate this skeleton in the generation process.


\section{Dialogue Response Generation}
\paragraph{Background} Dialogue systems can be grouped into two categories: chit-chat systems and task-oriented systems. While task-oriented dialogue systems are designed to accomplish specific user tasks such as air tickets booking, chit-chat dialogue systems aim at giving a meaningful and fluent response for any dialogue history in the open domain. Dialogue response generation in chit-chat dialogue system is challenging partly due to the diversity of possible responses to a single dialogue history (i.e., the \textit{one-to-many} problem). The dialogue history alone cannot decide a meaningful and specific response. Also, external knowledge that is not present in the dialogue history are often necessary for avoiding safe but boring responses. We focus on recent efforts tackling the challenges to develop chit-chat dialogue systems.

Most modern chit-chat dialogue systems can be categorized into two classes, namely, retrieval-based models and generation-based models. The retrieval-based models  \cite{ji2014information,hu2014convolutional} directly copy an existing response from curated dialogue corpora (i.e., the retrieval pool) when receiving a response request. The retrieved responses are often informative and grammatical as they are collected from real-world conversations and possibly post-edited by a human. However, such systems perform poorly when a given dialogue history is substantially different from those in the retrieval pool. On the other hand, the generation-based models \cite{shang-lu-li:2015:ACL-IJCNLP,vinyals2015neural,li-EtAl:2016:N16-11} generate a new utterance from scratch. Those generation-based models have better generalization capacity when handling unseen dialogue contexts. Nevertheless, the generated utterances are inclined to be dull and non-informative (e.g., ``I don't know'', ``I think so'', ``Me too'' etc.) \cite{li-EtAl:2016:N16-11}.

\paragraph{Shallow Integration} As discussed, retrieval-based models may give informative but inappropriate responses while generation-based models often do the opposite. It is desirable to combine the best of both worlds. Early work \cite{qiu-EtAl:2017:Short} attempts to re-rank the output from both models. For a deep integration, \newcite{song2016two} and \newcite{yang2019hybrid} extend the standard \textsc{Seq2Seq} encoder-decoder model \cite{bahdanau2014neural} with an extra encoder for encoding the retrieval result. The output of the extra encoder, along with the output from the original encoder for dialogue history, is used to feed the decoder. \newcite{weston-etal-2018-retrieve} use a single encoder that takes the concatenation of the original dialogue history and the retrieved as input. \newcite{wu2019response} note that the retrieved information should be used in awareness of the context difference, and further proposed to construct an edit vector by explicitly encoding the lexical differences between the input dialogue history and the retrieved dialogue history. \newcite{pandey2018exemplar} further propose to weight different training instances by context similarity.

\paragraph{Deep Integration} To prevent the inflow of erroneous information, \newcite{cai2019skeleton} propose a general framework that first extracts a skeleton from the retrieved response and then generates the response based on the extracted skeleton. This framework is also adopted for stylistic response generation \cite{su2021prototype}. \newcite{gupta-etal-2021-controlling} suggest to use the semantic structure of an exemplar response, instead of the tokens of the exemplar response, to guide generation. Despite their differences, a common issue is that the generation model easily learns to ignore the retrieved response entirely and collapses to a vanilla seq2seq model. This happens with improper training instances. Due to the one-to-many nature, it happens frequently that a retrieved response (extracted skeleton) is suitable for responding to the query, but inconsistent with the current target response.

Earlier studies \cite{weston-etal-2018-retrieve,wu2019response,cai2019skeleton} alleviate the above problems by putting hard constraints on the data (e.g., discarding data with low similarity of the retrieved response and the target response), which, however, greatly reduces the amount of usable data. \newcite{cai-etal-2019-retrieval} employ a random mechanism for generating the skeletons used for training, which extract skeletons from the corresponding responses with some deliberate disturbance. \newcite{paranjape2021hindsight} propose to model the retriever after the posterior distribution of retrieval given the input and the target output and train it jointly with the standard retriever and the generator by maximizing the evidence lower bound (ELBo) in expectation over retrieval.

\paragraph{Knowledge-Enhanced Generation} The aforementioned work demonstrates that retrieval-based dialogue systems can be used for building better generation-based models. In general, this is done by conditioning the generation on some retrieved responses. More traditionally, to infuse the response with external knowledge, the retrieval pool is not necessarily a dialogue corpus. In fact, knowledge-grounded dialogue response generation exploring different forms of knowledge such as knowledge bases and external documents \cite{dinan2018wizard,zhou2018dataset,lian2019learning,li2019incremental,qin2019conversing,wu2021controllable,zhang2021joint,komeili2021internet} has been actively explored.

\paragraph{Limitations} We note that there are three major limitations in existing work for dialogue response generation. First, current methods only use one retrieved response for generation. It can be more beneficial to combine multiple retrieval responses. However, this can be difficult due to the one-to-many nature of dialogue response generation. Second, current methods use universal relevance score for retrieval. It can be more effective if we can use more customized retrieval metric especially for controlled dialogue response generation (e.g., persona, emotion, etc).  Third, the retrieval pool of existing methods is limited to dialogue corpora (context-response pairs) or documents. It might be useful to enlarge the retrieval pool by including more corpora in other domains or in other modalities. As discussed, there leaves plenty of possible directions to explore in the future.

\section{Machine Translation}
Retrieval augmented translation originates from human translation scenarios~\cite{somers2003translation}. When translating $\hat{\boldsymbol{y}}$ from an input source sentence $\boldsymbol{x}$, a human translator typically involves a search engine to retrieve similar sentences $\{\langle \boldsymbol{x}^r, \boldsymbol{y}^r\rangle\}$  from a bilingual database. Such a technique called {\bf translation memory} is helpful to improve the translation quality and efficiency for human translators~\cite{dillon2006translators}. As the development of machine translation techniques, there is a surge of interests in improving machine translation models with translation memory. In the rest of this section, we will review translation memory for both statistical machine translation (SMT) and neural machine translation (NMT). 
\subsection{Translation Memory in SMT}

Generally, SMT includes three key components in a pipeline manner such as phrase table extraction, parameter tuning and decoding~\cite{koehn-etal-2003-statistical,chiang2007hierarchical}. As a result, many efforts have been made to make use of translation memory (TM) on top of each component. 

\paragraph{Constrained Decoding with TM}
Constrained decoding is the most straightforward way to integrating TM into SMT~\cite{smith2009ebmt,koehn2010convergence,zhechev2010seeding,ma2011consistent}. Its basic idea is to reuse the useful segments 
in $\boldsymbol{y}^r$ while translate other segments by SMT. Specifically, the approach consists of three steps: 1) identify the unmatched segments in both $\boldsymbol{x}^r$ and $\boldsymbol{x}$ through the edit-distance algorithm; 2)
identify the unmatched segments in $\boldsymbol{y}^r$, each of which is aligned to one unmatched segment in $\boldsymbol{x}^r$ by a word alignment algorithm; 3) decode each unmatched segment in $\boldsymbol{x}$ by SMT and then use the result to replace its corresponding unmatched segment in $\boldsymbol{y}^r$. \citet{li2016phrase} further extend this approach from sentence level to phrase level. 
The advantage in constrained decoding is that it does not require to change the translation model (including phrase table and parameters) and can be applied in a plug-and-play way. 
This approach is successful when $\boldsymbol{x}$ is highly similar to $\boldsymbol{x}^r$; otherwise its performance is degraded largely, because it explicitly isolates TM matching and SMT decoding and reuses the results in $\boldsymbol{x}^r$ or not in a deterministic way.  

\paragraph{Phrase Table Aggregation with TM}
There are also notable efforts to augment the phrase table for SMT by extracting translation rules from the retrieved bilingual sentences $\{\langle\boldsymbol{x}^r, \boldsymbol{y}^r\rangle\}$. Then they re-tune the parameters for the SMT model which makes use of translation knowledge from $\{\langle\boldsymbol{x}^r, \boldsymbol{y}^r\rangle\}$ in a implicit way when translating $\boldsymbol{x}$. For example, \citet{biccici2008dynamic,simard2009phrase} directly combine the extracted translation rules into the phrase table in a shallow combination way. They introduce an additional feature to indicate that whether translation rule is from $\{\langle\boldsymbol{x}^r, \boldsymbol{y}^r\rangle\}$ or not and then train all feature weights with MERT~\cite{och-2003-minimum}. One characteristic of these work is that a translation rule extracted from $\{\langle\boldsymbol{x}^r, \boldsymbol{y}^r\rangle\}$ which can not exactly match any segments in $\boldsymbol{x}$ is useless even if it may contain some useful words in its target side. To remedy this observation, \citet{wang2013integrating,wang2014dynamically} resort to a deep combination way to using the extracted translation rules. For each rule in the phrase table, it designs a generative model to reward the rules which are similar to those extracted from $\{\langle\boldsymbol{x}^r, \boldsymbol{y}^r\rangle\}$. Then this generative model is used as a feature in the log-linear based SMT model whose weight is tuned together with other features by MERT. In addition, \citet{li2014discriminative} employ a similar way to reward the rules but it relies on a discriminative model which is easy to integrate potential features from $\{\langle \boldsymbol{x}^r, \boldsymbol{y}^r\rangle\}$. 

\paragraph{Parameter Tuning with TM}
Unlike the above two research lines, 
\citet{liu2012locally,liu2014discriminative} make use of translation memory only in tuning parameters. To be specific, when translating an input sentence $\boldsymbol{x}$, they firstly retrieve many similar bilingual sentences $\{\langle \boldsymbol{x}^r, \boldsymbol{y}^r\rangle\}$, and then tune the  parameters on top of the retrieved sentences as well as a given development dataset in a sentence-wise manner, i.e., it performs an independent tuning for each input sentence. To improve the efficiency of each tuning step, it propose a local update on top of $\{\langle \boldsymbol{x}^r, \boldsymbol{y}^r\rangle\}$ from a baseline model.  


Despite the successes of translation memory in SMT, there are still some limitations for the above three kinds of methods. Firstly, all these methods employ fuzzy score for retrieval which is highly dependent on word matching and thus can not recall such examples which are similar in word semantics but different in surface form. Secondly, these methods integrate the retrieved examples into a module of SMT in the ways which can not make full use of the knowledge in retrieved examples. For example, the integration ways in the first two kinds (constrained decoding and phrase table aggregation) are heuristic and not optimized towards translation quality; the parameter tuning method fine-tunes few parameters for log-linear based SMT which are not enough to preserve sufficient knowledge from retrieved examples. 
Thirdly, since SMT performs in a pipeline manner, it is intractable to jointly optimize retrieval metrics as well as SMT models. Consequently, all these methods adopt an off-the-shelf metric for retrieval, leading to sub-optimal performance.

\subsection{Translation Memory in NMT}

Translation memory has been widely explored in Neural Machine Translation (NMT). Depending on when retrieval is involved, we can categorize previous works into two classes: 1) an NMT model leans how to cooperate with the retrieval model in the training phase; 2) an NMT model is only aware of the retrieved data in the inference phase.

\paragraph{Inference Phase} The key point of literature in this line is to reward some target words based on words in $\boldsymbol{y}^r$ in the inference process. Thus, a decision can be made based on both the distribution of  generation model and the additional reward of retrieval model. Some previous works propose to reward target words based on the sentence-level similarity between $\boldsymbol{x}$ and $\boldsymbol{x}^r$, and the word alignment between $\boldsymbol{x}^r$ and $\boldsymbol{y}^r$. Given the input sentence $\boldsymbol{x}$, \citet{zhang2018guiding} try to assign target words in $\hat{\boldsymbol{y}}$ with higher rewards, when they appear in $\boldsymbol{y}^r$ and the aligned source words  are in both $\boldsymbol{x}^r$ and $\boldsymbol{x}$. \citet{he2019word} follow a similar framework and consider the position information of those target words when rewarding. Those works reward the target words in an explicit way, however, the one-sentence-one-model approach \cite{li2016one, turchi2017continuous} propose to reward target word implicitly. For each testing input $\boldsymbol{x}$, their approach will first finetune the translation model on retrieved memory $\{\langle\boldsymbol{x}^r, \boldsymbol{y}^r\rangle\}$ and then translate $\boldsymbol{x}$.

Others try to reward target words based on token-level similarity score. Most works in this line are based on the dense retriever~\cite{khandelwal2020nearest}, e.g., faiss. \citet{khandelwal2020nearest} build a key-value datastore, where key $h(\boldsymbol{x}^r, \boldsymbol{y}^r_{<t})$ is the hidden state at each time step when translating $\boldsymbol{y}^r$ from $\boldsymbol{x}^r$, and value is its golden-truth target word $\boldsymbol{y}^r_{t}$. Therefore, in the inference time, they can use the $h(\boldsymbol{x}, \hat{\boldsymbol{y}}_{<t})$ as query and reward target words with similar hidden representations in the datastore. Although this method achieves significant performance gain, one drawback of it is the high latency. To address this issue, \citet{meng2021fast} use some heuristics, e.g., pre-filtering, to avoid searching on the entire datastore. The reward score of previous works is got from some non-parametric approaches, however, \citet{zheng2021adaptive} propose a light-weight network to learn the reward score. Since dense retrieval has the potential of cross-lingual retrieval, \citet{zheng2021non} use a similar approach to achieve unsupervised domain adaptation, where a main change is to create the datastore based on synthetic sources sentence and the real target sentences.


\paragraph{Training Phase} Different from those model-agnostic approaches, previous works in this line aim to train the generation model to learn how to cooperate with the retrieval model. It is also worth noting that most works in this line adopt the sentence-level retrieval, when integrating the retrieval information in the training process. To achieve its goal, \citet{bulte2019neural} and \citet{hossain2020simple} propose a data augmentation method to integrate the retrieved information, where $\boldsymbol{x}$ is concatenated with $\boldsymbol{y}^r$ before feeding into the model . Following the data augmentation approach, \citet{xu2020boosting} propose more matching methods to determine including which retrieved example in the source is better. 

There also exist some works that propose new architectures to integrate the retrieval information. Under the RNN-based framework, \citet{cao-xiong-2018-encoding} and \citet{gu2018search} use the gating and attention mechanism to incorporate the retrieved target sentences. When Transformer \cite{vaswani2017attention} becomes the backbone of NMT, some works also use additional transformer encoders to encode retrieved target sentences, and integrate them through attention mechanism \cite{bapna2019non, cao2019learning}. \citet{xia2019graph} represent the retrieved target sentences in a different data structure, i.e., a graph structure, and integrate it through attention mechanism. \citet{he2021fast} propose a light-weight method to encode the retrieved target sentences and leverage the alignment information to filter out irrelevant information. Different from previous works that rely on bilingual memories, \citet{cai-etal-2021-neural} propose a framework that can retrieve the most similar target sentence in a monolingual dataset, using a source sentence as query.

\paragraph{Limitations} In the section of SMT, we have showed some limitations of the retrieval augmented approaches. There also exist some limitations in the line of NMT. First, the information used for deriving reward scores is limited. The similarity between an input and retrieved examples is the primary feature to derive reward scores. However, some information, e.g., frequencies of words and context, may also be beneficial for integrating the translation memory. Second, it remains to be an open question that when should we use the retrieved information and when not. In the inference phase, approaches tend to integrate the translation memory excessively, e.g., at each time step, which not only reduces the translation efficiency but may also dampen the fluency of generated results.

\section{Other Tasks}

In addition to dialogue system and machine translation, retrieval-augmented generation techniques have shown to be beneficial in many other tasks. In the following, we highlight several key tasks that apply retrieval-augmented generation approaches.\footnote{Here, we focus on tasks other than question answering. We refer readers interested in QA to \citet{chen-yih-2020-open}.} 


\paragraph{Language Modelling} 
It has been shown that properly leveraging information from retrieval memory could improve the performance of large pre-trained language model. To build a more accurate language model, \citet{DBLP:conf/iclr/KhandelwalLJZL20} propose to incorporate a soft memory module into the system. Specifically, an index is built by caching the hidden states of the training corpus. Then, the language model accesses the index via k-NN search and displays a greatly improved performance. 
As another example, \citet{DBLP:journals/corr/abs-2002-08909} propose a new paradigm that applies retrieval-augmented technique into the pre-training of generative language model. During learning, they train a neural selector that dynamically samples a relevant text to guide the reconstruction of a corrupted input sequence. In this way, the pre-trained model delivers better results by explicitly grounding on the retrieval memory. \citet{DBLP:conf/nips/LewisGGAWZ20} combine language model pre-training with a paraphrasing approach. During learning, an input sequence to the model is first corrupted. In the meantime, a set of multi-lingual texts are retrieved based on which the model learns to reconstruct the original input sequence. 
Recently, \citet{DBLP:journals/corr/abs-2112-04426} propose \textsc{RETRO}, a large pre-trained language model enhanced with retrieved documents, and obtained comparable performances with GPT-3 using 25$\times$ fewer parameters.

\paragraph{Summarization} Text summarization is another research area that benefits from retrieval-augmented text generation. \citet{peng2019text} propose an adaptive decoding framework which first retrieves an exemplar document given the source document. Then, the summarization of the source document is derived through an adaptive generation process based on the retrieved template. Different from \citet{peng2019text}, \citet{DBLP:conf/acl/LiWLC18} and \citet{hossain2020simple} introduce an intermediate re-ranking stage into the generation pipeline. Specifically, before generating the document summary, the retrieval documents are first re-ranked based on their similarity scores with respect to the source document. Then, the document summarization is produced by re-writing the selected templates.

\paragraph{Paraphrase Generation} 
To address the lack of quality as well as diversity in the generation of paraphrases, \citet{kazemnejad-etal-2020-paraphrase} propose a generation framework which first retrieves a sentence that is similar to input sentence. Then, based on the retrieved sentence, a neural editor produces the resulting paraphrased sentence. \citet{DBLP:conf/acl/ChenTWG19} investigate a different aspect of paraphrasing, i.e. how to control the linguistic syntax displayed in the generated text. To achieve this goal, \citet{DBLP:conf/acl/ChenTWG19} propose to first extract a sentential exemplar that serves as the syntax template. A neural model then generates the paraphrase with desired linguistic syntax following the retrieved exemplar.



\paragraph{Text Style Transfer} To improve the quality of generated text, \citet{DBLP:conf/naacl/LiJHL18} propose a retrieval-augmented framework which first retrieves texts that are similar to the input based on lexical-level similarity. Then, the retrieved tokens that are irrelevant to the source are deleted, and the output is derived from the edited template. \citet{DBLP:conf/emnlp/XiaoPLWSC21} also adopte this framework by incorporating retrieval information from two sources (i.e. sparse and dense memories) and obtained an improved model performance.



\paragraph{Data-to-Text Generation} Recently, retrieval-augmented generation has been adapted to the task of data-to-text generation. To bridge the gap between the structured data and natural language text, \citet{DBLP:conf/emnlp/SuMBC21} propose a novel retrieval-augmented framework. Specifically, given the source data, a set of candidate texts are first retrieved from a large unlabelled corpus. Then, a neural selector is applied to measure the similarities between the source data and candidate texts, and extract a set of more fine-grained prototypes from the candidates. Lastly, a generation model takes the prototypes as input to produce the text that describes the given structured data.



While retrieval-augmented generation has been widely explored in the NLP community, we suggest that future research could extend this approach to tasks that involve data from multiple modalities. For instance, with recent advancements in image-text retrieval  \cite{DBLP:conf/icml/JiaYXCPPLSLD21,DBLP:conf/icml/RadfordKHRGASAM21}, the structural gap between images and texts is largely bridged. Some early studies \cite{DBLP:conf/iclr/0001C0USLZ20} have shown that information retrieved from images could improve the performance of neural machine translation model. Naturally, such methods could be extended to other multi-modal tasks, such as image captioning \cite{DBLP:conf/cvpr/KarpathyL15}. 
A similar idea could also be applied to tasks beyond images, such as 
speech-to-text transcription \cite{DBLP:journals/ftsig/GalesY07}.



\section{Future Directions}


Despite the current success of retrieval augmented text generation, there is still a long way to go as discussed in previous sections. We highlight some directions to facilitate the future research as follows:
\paragraph{Retrieval Sensitivity} The performance of retrieval augmented text generation is very sensitive to the retrieval quality, i.e., the similarity between the query and the retrieved examples. Currently, retrieval augmented text generation models perform well when the retrieved examples are very similar to the query. However, they are even worse than the generation models without retrieval when the retrieval examples are less similar. Therefore, it would be important to exploit new methods to address such an issue on similarity.
\paragraph{Retrieval Efficiency} Generally, if one enlarges the retrieval memory to some extent, it would be possible to retrieve an example which is very similar to the query.
    Unfortunately, the downside is that the overall inference for the retrieval augmented generation models is less efficient due the considerable retrieval overhead. In this sense, it is urgent to consider some methods to trade off the retrieval memory size and retrieval efficiency, for example, data compression for the retrieval memory.
\paragraph{Local vs. Global Optimization} Theoretically, it seems promising to jointly learn retrieval metrics and generation models. However, in practice, there is an essential gap about the retrieval metric between the training and inference phrases. In the training phase, the loss is locally back-propagated to only a few retrieved examples while in the inference phase the metric is globally conducted among all examples in the memory. It would be interesting to narrow such a gap when learning a better metric for generation tasks. 
    
 \paragraph{Multi-Modalities} With recent advancement in image-text retrieval, directly associating images with relevant text becomes possible. This urges researchers to investigate the possibility of retrieval-based text generation in tasks that involve data from different modalities. One typical task is image captioning. Beyond images, other tasks like speech-to-text transcription could potentially benefit from retrieval-based generation methods as well.
    
 \paragraph{Diverse \& Controllable Retrieval} Most of the existing approaches adopt a universal metric for retrieval, such as lexical similarities of sentences. Future work should explore how to use customized metrics for retrieval. This can be beneficial for more controlled text generation. For example, instances with emotions and styles may be more desirable in the personalized dialogue generation, parallel data that contains specific terminologies is more helpful in machine translation, and so on. On the other hand, using a universal metric for retrieval may lead to the lack of diversity of the retrieval results. Collecting a diverse set of retrieval results can improve the coverage of useful information. Thus, considering multiple different metrics for retrieval may lead to generation with higher quality in the future.

\section{Conclusion}
In this paper, we surveyed recent approaches for retrieval-augmented text generation. We reviewed and summarized the development of different components of retrieval-augmented text generation including retrieval metrics, retrieval sources, and integration paradigms. We gave in-depth discussions when retrieval-augmented text generation comes to different applications including dialogue response generation, machine translation, and other generation tasks. We also pointed out some future directions for retrieval-augmented text generation. 
\bibliography{anthology}

\begin{thebibliography}{83}
\expandafter\ifx\csname natexlab\endcsname\relax\def\natexlab#1{#1}\fi

\bibitem[{Bahdanau et~al.(2014)Bahdanau, Cho, and Bengio}]{bahdanau2014neural}
Dzmitry Bahdanau, Kyunghyun Cho, and Yoshua Bengio. 2014.
\newblock Neural machine translation by jointly learning to align and
  translate.
\newblock \emph{arXiv preprint arXiv:1409.0473}.

\bibitem[{Bapna and Firat(2019)}]{bapna2019non}
Ankur Bapna and Orhan Firat. 2019.
\newblock Non-parametric adaptation for neural machine translation.
\newblock In \emph{Proceedings of the 2019 Conference of the North American
  Chapter of the Association for Computational Linguistics: Human Language
  Technologies, Volume 1 (Long and Short Papers)}, pages 1921--1931.

\bibitem[{Bi{\c{c}}ici and Dymetman(2008)}]{biccici2008dynamic}
Ergun Bi{\c{c}}ici and Marc Dymetman. 2008.
\newblock Dynamic translation memory: Using statistical machine translation to
  improve translation memory fuzzy matches.
\newblock In \emph{International Conference on Intelligent Text Processing and
  Computational Linguistics}, pages 454--465. Springer.

\bibitem[{Borgeaud et~al.(2021)Borgeaud, Mensch, Hoffmann, Cai, Rutherford,
  Millican, van~den Driessche, Lespiau, Damoc, Clark, de~Las~Casas, Guy,
  Menick, Ring, Hennigan, Huang, Maggiore, Jones, Cassirer, Brock, Paganini,
  Irving, Vinyals, Osindero, Simonyan, Rae, Elsen, and
  Sifre}]{DBLP:journals/corr/abs-2112-04426}
Sebastian Borgeaud, Arthur Mensch, Jordan Hoffmann, Trevor Cai, Eliza
  Rutherford, Katie Millican, George van~den Driessche, Jean{-}Baptiste
  Lespiau, Bogdan Damoc, Aidan Clark, Diego de~Las~Casas, Aurelia Guy, Jacob
  Menick, Roman Ring, Tom Hennigan, Saffron Huang, Loren Maggiore, Chris Jones,
  Albin Cassirer, Andy Brock, Michela Paganini, Geoffrey Irving, Oriol Vinyals,
  Simon Osindero, Karen Simonyan, Jack~W. Rae, Erich Elsen, and Laurent Sifre.
  2021.
\newblock \href {http://arxiv.org/abs/2112.04426} {Improving language models by
  retrieving from trillions of tokens}.
\newblock \emph{CoRR}, abs/2112.04426.

\bibitem[{Bulte and Tezcan(2019)}]{bulte2019neural}
Bram Bulte and Arda Tezcan. 2019.
\newblock Neural fuzzy repair: Integrating fuzzy matches into neural machine
  translation.
\newblock In \emph{Proceedings of the 57th Annual Meeting of the Association
  for Computational Linguistics}, pages 1800--1809.

\bibitem[{Cai et~al.(2019{\natexlab{a}})Cai, Wang, Bi, Tu, Liu, Lam, and
  Shi}]{cai2019skeleton}
Deng Cai, Yan Wang, Wei Bi, Zhaopeng Tu, Xiaojiang Liu, Wai Lam, and Shuming
  Shi. 2019{\natexlab{a}}.
\newblock Skeleton-to-response: Dialogue generation guided by retrieval memory.
\newblock In \emph{Proceedings of the 2019 Conference of the North {A}merican
  Chapter of the Association for Computational Linguistics: Human Language
  Technologies, Volume 1 (Long and Short Papers)}, pages 1219--1228.

\bibitem[{Cai et~al.(2019{\natexlab{b}})Cai, Wang, Bi, Tu, Liu, and
  Shi}]{cai-etal-2019-retrieval}
Deng Cai, Yan Wang, Wei Bi, Zhaopeng Tu, Xiaojiang Liu, and Shuming Shi.
  2019{\natexlab{b}}.
\newblock Retrieval-guided dialogue response generation via a
  matching-to-generation framework.
\newblock In \emph{Proceedings of the 2019 Conference on Empirical Methods in
  Natural Language Processing and the 9th International Joint Conference on
  Natural Language Processing (EMNLP-IJCNLP)}, pages 1866--1875.

\bibitem[{Cai et~al.(2021)Cai, Wang, Li, Lam, and Liu}]{cai-etal-2021-neural}
Deng Cai, Yan Wang, Huayang Li, Wai Lam, and Lemao Liu. 2021.
\newblock \href {https://doi.org/10.18653/v1/2021.acl-long.567} {Neural machine
  translation with monolingual translation memory}.
\newblock In \emph{Proceedings of the 59th Annual Meeting of the Association
  for Computational Linguistics and the 11th International Joint Conference on
  Natural Language Processing (Volume 1: Long Papers)}, pages 7307--7318,
  Online. Association for Computational Linguistics.

\bibitem[{Cao et~al.(2019)Cao, Kuang, and Xiong}]{cao2019learning}
Qian Cao, Shaohui Kuang, and Deyi Xiong. 2019.
\newblock Learning to reuse translations: Guiding neural machine translation
  with examples.
\newblock \emph{arXiv preprint arXiv:1911.10732}.

\bibitem[{Cao and Xiong(2018)}]{cao-xiong-2018-encoding}
Qian Cao and Deyi Xiong. 2018.
\newblock Encoding gated translation memory into neural machine translation.
\newblock In \emph{Proceedings of the 2018 Conference on Empirical Methods in
  Natural Language Processing}, pages 3042--3047.

\bibitem[{Cao et~al.(2018)Cao, Li, Li, and Wei}]{DBLP:conf/acl/LiWLC18}
Ziqiang Cao, Wenjie Li, Sujian Li, and Furu Wei. 2018.
\newblock \href {https://doi.org/10.18653/v1/P18-1015} {Retrieve, rerank and
  rewrite: Soft template based neural summarization}.
\newblock In \emph{Proceedings of the 56th Annual Meeting of the Association
  for Computational Linguistics, {ACL} 2018, Melbourne, Australia, July 15-20,
  2018, Volume 1: Long Papers}, pages 152--161. Association for Computational
  Linguistics.

\bibitem[{Chen and Yih(2020)}]{chen-yih-2020-open}
Danqi Chen and Wen-tau Yih. 2020.
\newblock \href {https://doi.org/10.18653/v1/2020.acl-tutorials.8} {Open-domain
  question answering}.
\newblock In \emph{Proceedings of the 58th Annual Meeting of the Association
  for Computational Linguistics: Tutorial Abstracts}, pages 34--37, Online.
  Association for Computational Linguistics.

\bibitem[{Chen et~al.(2019)Chen, Tang, Wiseman, and
  Gimpel}]{DBLP:conf/acl/ChenTWG19}
Mingda Chen, Qingming Tang, Sam Wiseman, and Kevin Gimpel. 2019.
\newblock \href {https://doi.org/10.18653/v1/p19-1599} {Controllable paraphrase
  generation with a syntactic exemplar}.
\newblock In \emph{Proceedings of the 57th Conference of the Association for
  Computational Linguistics, {ACL} 2019, Florence, Italy, July 28- August 2,
  2019, Volume 1: Long Papers}, pages 5972--5984. Association for Computational
  Linguistics.

\bibitem[{Chiang(2007)}]{chiang2007hierarchical}
David Chiang. 2007.
\newblock Hierarchical phrase-based translation.
\newblock \emph{computational linguistics}, 33(2):201--228.

\bibitem[{Dillon and Fraser(2006)}]{dillon2006translators}
Sarah Dillon and Janet Fraser. 2006.
\newblock Translators and tm: An investigation of translators’ perceptions of
  translation memory adoption.
\newblock \emph{Machine Translation}, 20(2):67--79.

\bibitem[{Dinan et~al.(2018)Dinan, Roller, Shuster, Fan, Auli, and
  Weston}]{dinan2018wizard}
Emily Dinan, Stephen Roller, Kurt Shuster, Angela Fan, Michael Auli, and Jason
  Weston. 2018.
\newblock Wizard of wikipedia: Knowledge-powered conversational agents.
\newblock \emph{arXiv preprint arXiv:1811.01241}.

\bibitem[{Gales and Young(2007)}]{DBLP:journals/ftsig/GalesY07}
Mark J.~F. Gales and Steve~J. Young. 2007.
\newblock \href {https://doi.org/10.1561/2000000004} {The application of hidden
  markov models in speech recognition}.
\newblock \emph{Found. Trends Signal Process.}, 1(3):195--304.

\bibitem[{Gu et~al.(2018)Gu, Wang, Cho, and Li}]{gu2018search}
Jiatao Gu, Yong Wang, Kyunghyun Cho, and Victor~OK Li. 2018.
\newblock Search engine guided neural machine translation.
\newblock In \emph{Proceedings of the AAAI Conference on Artificial
  Intelligence}, volume~32.

\bibitem[{Gupta et~al.(2021)Gupta, Bigham, Tsvetkov, and
  Pavel}]{gupta-etal-2021-controlling}
Prakhar Gupta, Jeffrey Bigham, Yulia Tsvetkov, and Amy Pavel. 2021.
\newblock \href {https://doi.org/10.18653/v1/2021.naacl-main.240} {Controlling
  dialogue generation with semantic exemplars}.
\newblock In \emph{Proceedings of the 2021 Conference of the North American
  Chapter of the Association for Computational Linguistics: Human Language
  Technologies}, pages 3018--3029, Online. Association for Computational
  Linguistics.

\bibitem[{Guu et~al.(2020)Guu, Lee, Tung, Pasupat, and
  Chang}]{DBLP:journals/corr/abs-2002-08909}
Kelvin Guu, Kenton Lee, Zora Tung, Panupong Pasupat, and Ming{-}Wei Chang.
  2020.
\newblock \href {http://arxiv.org/abs/2002.08909} {{REALM:} retrieval-augmented
  language model pre-training}.
\newblock \emph{CoRR}, abs/2002.08909.

\bibitem[{Hashimoto et~al.(2018)Hashimoto, Guu, Oren, and
  Liang}]{hashimoto2018retrieve}
Tatsunori~B Hashimoto, Kelvin Guu, Yonatan Oren, and Percy~S Liang. 2018.
\newblock A retrieve-and-edit framework for predicting structured outputs.
\newblock In \emph{Advances in Neural Information Processing Systems}, pages
  10052--10062.

\bibitem[{He et~al.(2021)He, Huang, Cui, Li, and Liu}]{he2021fast}
Qiuxiang He, Guoping Huang, Qu~Cui, Li~Li, and Lemao Liu. 2021.
\newblock Fast and accurate neural machine translation with translation memory.
\newblock In \emph{Proceedings of the 59th Annual Meeting of the Association
  for Computational Linguistics and the 11th International Joint Conference on
  Natural Language Processing (Volume 1: Long Papers)}, pages 3170--3180.

\bibitem[{He et~al.(2019)He, Huang, Liu, and Li}]{he2019word}
Qiuxiang He, Guoping Huang, Lemao Liu, and Li~Li. 2019.
\newblock Word position aware translation memory for neural machine
  translation.
\newblock In \emph{CCF International Conference on Natural Language Processing
  and Chinese Computing}, pages 367--379. Springer.

\bibitem[{Hossain et~al.(2020)Hossain, Ghazvininejad, and
  Zettlemoyer}]{hossain2020simple}
Nabil Hossain, Marjan Ghazvininejad, and Luke Zettlemoyer. 2020.
\newblock Simple and effective retrieve-edit-rerank text generation.
\newblock In \emph{Proceedings of the 58th Annual Meeting of the Association
  for Computational Linguistics}, pages 2532--2538.

\bibitem[{Hu et~al.(2014)Hu, Lu, Li, and Chen}]{hu2014convolutional}
Baotian Hu, Zhengdong Lu, Hang Li, and Qingcai Chen. 2014.
\newblock Convolutional neural network architectures for matching natural
  language sentences.
\newblock In \emph{NIPS}, pages 2042--2050.

\bibitem[{Ji et~al.(2014)Ji, Lu, and Li}]{ji2014information}
Zongcheng Ji, Zhengdong Lu, and Hang Li. 2014.
\newblock An information retrieval approach to short text conversation.
\newblock \emph{arXiv preprint arXiv:1408.6988}.

\bibitem[{Jia et~al.(2021)Jia, Yang, Xia, Chen, Parekh, Pham, Le, Sung, Li, and
  Duerig}]{DBLP:conf/icml/JiaYXCPPLSLD21}
Chao Jia, Yinfei Yang, Ye~Xia, Yi{-}Ting Chen, Zarana Parekh, Hieu Pham,
  Quoc~V. Le, Yun{-}Hsuan Sung, Zhen Li, and Tom Duerig. 2021.
\newblock \href {http://proceedings.mlr.press/v139/jia21b.html} {Scaling up
  visual and vision-language representation learning with noisy text
  supervision}.
\newblock In \emph{Proceedings of the 38th International Conference on Machine
  Learning, {ICML} 2021, 18-24 July 2021, Virtual Event}, volume 139 of
  \emph{Proceedings of Machine Learning Research}, pages 4904--4916. {PMLR}.

\bibitem[{Karpathy and Li(2015)}]{DBLP:conf/cvpr/KarpathyL15}
Andrej Karpathy and Fei-Fei Li. 2015.
\newblock \href {https://doi.org/10.1109/CVPR.2015.7298932} {Deep
  visual-semantic alignments for generating image descriptions}.
\newblock In \emph{{IEEE} Conference on Computer Vision and Pattern
  Recognition, {CVPR} 2015, Boston, MA, USA, June 7-12, 2015}, pages
  3128--3137. {IEEE} Computer Society.

\bibitem[{Kazemnejad et~al.(2020)Kazemnejad, Salehi, and
  Soleymani~Baghshah}]{kazemnejad-etal-2020-paraphrase}
Amirhossein Kazemnejad, Mohammadreza Salehi, and Mahdieh Soleymani~Baghshah.
  2020.
\newblock \href {https://doi.org/10.18653/v1/2020.acl-main.535} {Paraphrase
  generation by learning how to edit from samples}.
\newblock In \emph{Proceedings of the 58th Annual Meeting of the Association
  for Computational Linguistics}, pages 6010--6021, Online. Association for
  Computational Linguistics.

\bibitem[{Khandelwal et~al.(2020{\natexlab{a}})Khandelwal, Fan, Jurafsky,
  Zettlemoyer, and Lewis}]{khandelwal2020nearest}
Urvashi Khandelwal, Angela Fan, Dan Jurafsky, Luke Zettlemoyer, and Mike Lewis.
  2020{\natexlab{a}}.
\newblock Nearest neighbor machine translation.
\newblock \emph{arXiv preprint arXiv:2010.00710}.

\bibitem[{Khandelwal et~al.(2020{\natexlab{b}})Khandelwal, Levy, Jurafsky,
  Zettlemoyer, and Lewis}]{DBLP:conf/iclr/KhandelwalLJZL20}
Urvashi Khandelwal, Omer Levy, Dan Jurafsky, Luke Zettlemoyer, and Mike Lewis.
  2020{\natexlab{b}}.
\newblock \href {https://openreview.net/forum?id=HklBjCEKvH} {Generalization
  through memorization: Nearest neighbor language models}.
\newblock In \emph{8th International Conference on Learning Representations,
  {ICLR} 2020, Addis Ababa, Ethiopia, April 26-30, 2020}. OpenReview.net.

\bibitem[{Koehn et~al.(2003)Koehn, Och, and
  Marcu}]{koehn-etal-2003-statistical}
Philipp Koehn, Franz~J. Och, and Daniel Marcu. 2003.
\newblock \href {https://aclanthology.org/N03-1017} {Statistical phrase-based
  translation}.
\newblock In \emph{Proceedings of the 2003 Human Language Technology Conference
  of the North {A}merican Chapter of the Association for Computational
  Linguistics}, pages 127--133.

\bibitem[{Koehn and Senellart(2010)}]{koehn2010convergence}
Philipp Koehn and Jean Senellart. 2010.
\newblock Convergence of translation memory and statistical machine
  translation.
\newblock In \emph{Proceedings of AMTA Workshop on MT Research and the
  Translation Industry}, pages 21--31.

\bibitem[{Komeili et~al.(2021)Komeili, Shuster, and
  Weston}]{komeili2021internet}
Mojtaba Komeili, Kurt Shuster, and Jason Weston. 2021.
\newblock Internet-augmented dialogue generation.
\newblock \emph{arXiv preprint arXiv:2107.07566}.

\bibitem[{Lee et~al.(2019)Lee, Chang, and Toutanova}]{lee2019latent}
Kenton Lee, Ming-Wei Chang, and Kristina Toutanova. 2019.
\newblock Latent retrieval for weakly supervised open domain question
  answering.
\newblock \emph{arXiv preprint arXiv:1906.00300}.

\bibitem[{Lewis et~al.(2020{\natexlab{a}})Lewis, Ghazvininejad, Ghosh,
  Aghajanyan, Wang, and Zettlemoyer}]{DBLP:conf/nips/LewisGGAWZ20}
Mike Lewis, Marjan Ghazvininejad, Gargi Ghosh, Armen Aghajanyan, Sida Wang, and
  Luke Zettlemoyer. 2020{\natexlab{a}}.
\newblock \href
  {https://proceedings.neurips.cc/paper/2020/hash/d6f1dd034aabde7657e6680444ceff62-Abstract.html}
  {Pre-training via paraphrasing}.
\newblock In \emph{Advances in Neural Information Processing Systems 33: Annual
  Conference on Neural Information Processing Systems 2020, NeurIPS 2020,
  December 6-12, 2020, virtual}.

\bibitem[{Lewis et~al.(2020{\natexlab{b}})Lewis, Perez, Piktus, Petroni,
  Karpukhin, Goyal, K{\"u}ttler, Lewis, Yih, Rockt{\"a}schel
  et~al.}]{lewis2020retrieval}
Patrick Lewis, Ethan Perez, Aleksandra Piktus, Fabio Petroni, Vladimir
  Karpukhin, Naman Goyal, Heinrich K{\"u}ttler, Mike Lewis, Wen-tau Yih, Tim
  Rockt{\"a}schel, et~al. 2020{\natexlab{b}}.
\newblock Retrieval-augmented generation for knowledge-intensive nlp tasks.
\newblock \emph{arXiv preprint arXiv:2005.11401}.

\bibitem[{Li et~al.(2016{\natexlab{a}})Li, Galley, Brockett, Gao, and
  Dolan}]{li-EtAl:2016:N16-11}
Jiwei Li, Michel Galley, Chris Brockett, Jianfeng Gao, and Bill Dolan.
  2016{\natexlab{a}}.
\newblock A diversity-promoting objective function for neural conversation
  models.
\newblock In \emph{NAACL}, pages 110--119.

\bibitem[{Li et~al.(2018)Li, Jia, He, and Liang}]{DBLP:conf/naacl/LiJHL18}
Juncen Li, Robin Jia, He~He, and Percy Liang. 2018.
\newblock \href {https://doi.org/10.18653/v1/n18-1169} {Delete, retrieve,
  generate: a simple approach to sentiment and style transfer}.
\newblock In \emph{Proceedings of the 2018 Conference of the North American
  Chapter of the Association for Computational Linguistics: Human Language
  Technologies, {NAACL-HLT} 2018, New Orleans, Louisiana, USA, June 1-6, 2018,
  Volume 1 (Long Papers)}, pages 1865--1874. Association for Computational
  Linguistics.

\bibitem[{Li et~al.(2014)Li, Way, and Liu}]{li2014discriminative}
Liangyou Li, Andy Way, and Qun Liu. 2014.
\newblock A discriminative framework of integrating translation memory features
  into smt.
\newblock In \emph{Proceedings of the 11th Conference of the Association for
  Machine Translation in the Americas}, volume~1, pages 249--260.

\bibitem[{Li et~al.(2016{\natexlab{b}})Li, Way, and Liu}]{li2016phrase}
Liangyou Li, Andy Way, and Qun Liu. 2016{\natexlab{b}}.
\newblock Phrase-level combination of smt and tm using constrained word
  lattice.
\newblock Association for Computational Linguistics (ACL).

\bibitem[{Li et~al.(2016{\natexlab{c}})Li, Zhang, and Zong}]{li2016one}
Xiaoqing Li, Jiajun Zhang, and Chengqing Zong. 2016{\natexlab{c}}.
\newblock One sentence one model for neural machine translation.
\newblock \emph{arXiv preprint arXiv:1609.06490}.

\bibitem[{Li et~al.(2019)Li, Niu, Meng, Feng, Li, and Zhou}]{li2019incremental}
Zekang Li, Cheng Niu, Fandong Meng, Yang Feng, Qian Li, and Jie Zhou. 2019.
\newblock Incremental transformer with deliberation decoder for document
  grounded conversations.
\newblock In \emph{Proceedings of the 57th Annual Meeting of the Association
  for Computational Linguistics}, pages 12--21.

\bibitem[{Lian et~al.(2019)Lian, Xie, Wang, Peng, and Wu}]{lian2019learning}
Rongzhong Lian, Min Xie, Fan Wang, Jinhua Peng, and Hua Wu. 2019.
\newblock Learning to select knowledge for response generation in dialog
  systems.
\newblock \emph{arXiv preprint arXiv:1902.04911}.

\bibitem[{Liu et~al.(2012)Liu, Cao, Watanabe, Zhao, Yu, and
  Zhu}]{liu2012locally}
Lemao Liu, Hailong Cao, Taro Watanabe, Tiejun Zhao, Mo~Yu, and Conghui Zhu.
  2012.
\newblock Locally training the log-linear model for smt.
\newblock In \emph{Proceedings of the 2012 Joint Conference on Empirical
  Methods in Natural Language Processing and Computational Natural Language
  Learning}, pages 402--411.

\bibitem[{Liu et~al.(2014)Liu, Zhao, Watanabe, Cao, and
  Zhu}]{liu2014discriminative}
Lemao Liu, Tiejun Zhao, Taro Watanabe, Hailong Cao, and Conghui Zhu. 2014.
\newblock Discriminative training for log-linear based smt: Global or local
  methods.
\newblock \emph{ACM Transactions on Asian Language Information Processing
  (TALIP)}, 13(4):1--25.

\bibitem[{Ma et~al.(2011)Ma, He, Way, and van Genabith}]{ma2011consistent}
Yanjun Ma, Yifan He, Andy Way, and Josef van Genabith. 2011.
\newblock Consistent translation using discriminative learning-a translation
  memory-inspired approach.
\newblock In \emph{Proceedings of the 49th Annual Meeting of the Association
  for Computational Linguistics: Human Language Technologies}, pages
  1239--1248.

\bibitem[{Meng et~al.(2021)Meng, Li, Zheng, Wu, Sun, Zhang, and
  Li}]{meng2021fast}
Yuxian Meng, Xiaoya Li, Xiayu Zheng, Fei Wu, Xiaofei Sun, Tianwei Zhang, and
  Jiwei Li. 2021.
\newblock Fast nearest neighbor machine translation.
\newblock \emph{arXiv preprint arXiv:2105.14528}.

\bibitem[{Och(2003)}]{och-2003-minimum}
Franz~Josef Och. 2003.
\newblock \href {https://doi.org/10.3115/1075096.1075117} {Minimum error rate
  training in statistical machine translation}.
\newblock In \emph{Proceedings of the 41st Annual Meeting of the Association
  for Computational Linguistics}, pages 160--167, Sapporo, Japan. Association
  for Computational Linguistics.

\bibitem[{Pandey et~al.(2018)Pandey, Contractor, Kumar, and
  Joshi}]{pandey2018exemplar}
Gaurav Pandey, Danish Contractor, Vineet Kumar, and Sachindra Joshi. 2018.
\newblock Exemplar encoder-decoder for neural conversation generation.
\newblock In \emph{ACL}, pages 1329--1338.

\bibitem[{Paranjape et~al.(2021)Paranjape, Khattab, Potts, Zaharia, and
  Manning}]{paranjape2021hindsight}
Ashwin Paranjape, Omar Khattab, Christopher Potts, Matei Zaharia, and
  Christopher~D Manning. 2021.
\newblock Hindsight: Posterior-guided training of retrievers for improved
  open-ended generation.
\newblock \emph{arXiv preprint arXiv:2110.07752}.

\bibitem[{Peng et~al.(2019)Peng, Parikh, Faruqui, Dhingra, and
  Dipanjan}]{peng2019text}
Hao Peng, Ankur~P. Parikh, Manaal Faruqui, Bhuwan Dhingra, and Das Dipanjan.
  2019.
\newblock Text generation with exemplar-based adaptive decoding.
\newblock In \emph{Proceedings of the Conference of the North American Chapter
  of the Association for Computational Linguistics: Human Language
  Technologies}.

\bibitem[{Qin et~al.(2019)Qin, Galley, Brockett, Liu, Gao, Dolan, Choi, and
  Gao}]{qin2019conversing}
Lianhui Qin, Michel Galley, Chris Brockett, Xiaodong Liu, Xiang Gao, William~B
  Dolan, Yejin Choi, and Jianfeng Gao. 2019.
\newblock Conversing by reading: Contentful neural conversation with on-demand
  machine reading.
\newblock In \emph{Proceedings of the 57th Annual Meeting of the Association
  for Computational Linguistics}, pages 5427--5436.

\bibitem[{Qiu et~al.(2017)Qiu, Li, Wang, Gao, Chen, Zhao, Chen, Huang, and
  Chu}]{qiu-EtAl:2017:Short}
Minghui Qiu, Feng-Lin Li, Siyu Wang, Xing Gao, Yan Chen, Weipeng Zhao, Haiqing
  Chen, Jun Huang, and Wei Chu. 2017.
\newblock Alime chat: A sequence to sequence and rerank based chatbot engine.
\newblock In \emph{ACL}, pages 498--503.

\bibitem[{Radford et~al.(2021)Radford, Kim, Hallacy, Ramesh, Goh, Agarwal,
  Sastry, Askell, Mishkin, Clark, Krueger, and
  Sutskever}]{DBLP:conf/icml/RadfordKHRGASAM21}
Alec Radford, Jong~Wook Kim, Chris Hallacy, Aditya Ramesh, Gabriel Goh,
  Sandhini Agarwal, Girish Sastry, Amanda Askell, Pamela Mishkin, Jack Clark,
  Gretchen Krueger, and Ilya Sutskever. 2021.
\newblock \href {http://proceedings.mlr.press/v139/radford21a.html} {Learning
  transferable visual models from natural language supervision}.
\newblock In \emph{Proceedings of the 38th International Conference on Machine
  Learning, {ICML} 2021, 18-24 July 2021, Virtual Event}, volume 139 of
  \emph{Proceedings of Machine Learning Research}, pages 8748--8763. {PMLR}.

\bibitem[{Robertson and Zaragoza(2009)}]{robertson2009probabilistic}
Stephen Robertson and Hugo Zaragoza. 2009.
\newblock \emph{The probabilistic relevance framework: BM25 and beyond}.
\newblock Now Publishers Inc.

\bibitem[{Shang et~al.(2015)Shang, Lu, and Li}]{shang-lu-li:2015:ACL-IJCNLP}
Lifeng Shang, Zhengdong Lu, and Hang Li. 2015.
\newblock Neural responding machine for short-text conversation.
\newblock In \emph{ACL}, pages 1577--1586.

\bibitem[{Simard and Isabelle(2009)}]{simard2009phrase}
Michel Simard and Pierre Isabelle. 2009.
\newblock Phrase-based machine translation in a computer-assisted translation
  environment.
\newblock \emph{Proceedings of the Twelfth Machine Translation Summit (MT
  Summit XII)}, pages 120--127.

\bibitem[{Smith and Clark(2009)}]{smith2009ebmt}
James Smith and Stephen Clark. 2009.
\newblock Ebmt for smt: a new ebmt-smt hybrid.
\newblock In \emph{Proceedings of the 3rd International Workshop on
  Example-Based Machine Translation}, pages 3--10. Citeseer.

\bibitem[{Somers(2003)}]{somers2003translation}
Harold Somers. 2003.
\newblock Translation memory systems.
\newblock \emph{Benjamins Translation Library}, 35:31--48.

\bibitem[{Song et~al.(2016)Song, Yan, Li, Zhao, and Zhang}]{song2016two}
Yiping Song, Rui Yan, Xiang Li, Dongyan Zhao, and Ming Zhang. 2016.
\newblock Two are better than one: An ensemble of retrieval-and
  generation-based dialog systems.
\newblock \emph{arXiv preprint arXiv:1610.07149}.

\bibitem[{Su et~al.(2021{\natexlab{a}})Su, Meng, Baker, and
  Collier}]{DBLP:conf/emnlp/SuMBC21}
Yixuan Su, Zaiqiao Meng, Simon Baker, and Nigel Collier. 2021{\natexlab{a}}.
\newblock \href {https://aclanthology.org/2021.findings-emnlp.77} {Few-shot
  table-to-text generation with prototype memory}.
\newblock In \emph{Findings of the Association for Computational Linguistics:
  {EMNLP} 2021, Virtual Event / Punta Cana, Dominican Republic, 16-20 November,
  2021}, pages 910--917. Association for Computational Linguistics.

\bibitem[{Su et~al.(2021{\natexlab{b}})Su, Vandyke, Baker, Wang, and
  Collier}]{su-etal-2021-keep}
Yixuan Su, David Vandyke, Simon Baker, Yan Wang, and Nigel Collier.
  2021{\natexlab{b}}.
\newblock \href {https://doi.org/10.18653/v1/2021.findings-acl.50} {Keep the
  primary, rewrite the secondary: A two-stage approach for paraphrase
  generation}.
\newblock In \emph{Findings of the Association for Computational Linguistics:
  ACL-IJCNLP 2021}, pages 560--569, Online. Association for Computational
  Linguistics.

\bibitem[{Su et~al.(2021{\natexlab{c}})Su, Wang, Cai, Baker, Korhonen, and
  Collier}]{su2021prototype}
Yixuan Su, Yan Wang, Deng Cai, Simon Baker, Anna Korhonen, and Nigel Collier.
  2021{\natexlab{c}}.
\newblock \href {https://doi.org/10.1109/TASLP.2021.3087948}
  {{PROTOTYPE-TO-STYLE:} dialogue generation with style-aware editing on
  retrieval memory}.
\newblock \emph{{IEEE} {ACM} Trans. Audio Speech Lang. Process.},
  29:2152--2161.

\bibitem[{Turchi et~al.(2017)Turchi, Negri, Farajian, and
  Federico}]{turchi2017continuous}
Marco Turchi, Matteo Negri, M~Farajian, and Marcello Federico. 2017.
\newblock Continuous learning from human post-edits for neural machine
  translation.

\bibitem[{Vaswani et~al.(2017)Vaswani, Shazeer, Parmar, Uszkoreit, Jones,
  Gomez, Kaiser, and Polosukhin}]{vaswani2017attention}
Ashish Vaswani, Noam Shazeer, Niki Parmar, Jakob Uszkoreit, Llion Jones,
  Aidan~N Gomez, {\L}ukasz Kaiser, and Illia Polosukhin. 2017.
\newblock Attention is all you need.
\newblock In \emph{Advances in neural information processing systems}, pages
  5998--6008.

\bibitem[{Vinyals and Le(2015)}]{vinyals2015neural}
Oriol Vinyals and Quoc Le. 2015.
\newblock A neural conversational model.
\newblock In \emph{ICML (Deep Learning Workshop)}.

\bibitem[{Wang et~al.(2013)Wang, Zong, and Su}]{wang2013integrating}
Kun Wang, Chengqing Zong, and Keh-Yih Su. 2013.
\newblock Integrating translation memory into phrase-based machine translation
  during decoding.
\newblock In \emph{Proceedings of the 51st Annual Meeting of the Association
  for Computational Linguistics (Volume 1: Long Papers)}, pages 11--21.

\bibitem[{Wang et~al.(2014)Wang, Zong, and Su}]{wang2014dynamically}
Kun Wang, Chengqing Zong, and Keh-Yih Su. 2014.
\newblock Dynamically integrating cross-domain translation memory into
  phrase-based machine translation during decoding.
\newblock In \emph{Proceedings of COLING 2014, the 25th International
  Conference on Computational Linguistics: Technical Papers}, pages 398--408.

\bibitem[{Weston et~al.(2018)Weston, Dinan, and
  Miller}]{weston-etal-2018-retrieve}
Jason Weston, Emily Dinan, and Alexander Miller. 2018.
\newblock Retrieve and refine: Improved sequence generation models for
  dialogue.
\newblock In \emph{Proceedings of the 2018 {EMNLP} Workshop {SCAI}: The 2nd
  International Workshop on Search-Oriented Conversational {AI}}, pages 87--92.

\bibitem[{Wu et~al.(2019)Wu, Wei, Huang, Wang, Li, and Zhou}]{wu2019response}
Yu~Wu, Furu Wei, Shaohan Huang, Yunli Wang, Zhoujun Li, and Ming Zhou. 2019.
\newblock Response generation by context-aware prototype editing.
\newblock In \emph{Proceedings of the AAAI Conference on Artificial
  Intelligence}, volume~33, pages 7281--7288.

\bibitem[{Wu et~al.(2021)Wu, Galley, Brockett, Zhang, Gao, Quirk,
  Koncel-Kedziorski, Gao, Hajishirzi, Ostendorf et~al.}]{wu2021controllable}
Zeqiu Wu, Michel Galley, Chris Brockett, Yizhe Zhang, Xiang Gao, Chris Quirk,
  Rik Koncel-Kedziorski, Jianfeng Gao, Hannaneh Hajishirzi, Mari Ostendorf,
  et~al. 2021.
\newblock A controllable model of grounded response generation.
\newblock In \emph{Proceedings of the AAAI Conference on Artificial
  Intelligence}, volume~35, pages 14085--14093.

\bibitem[{Xia et~al.(2019)Xia, Huang, Liu, and Shi}]{xia2019graph}
Mengzhou Xia, Guoping Huang, Lemao Liu, and Shuming Shi. 2019.
\newblock Graph based translation memory for neural machine translation.
\newblock In \emph{Proceedings of the AAAI Conference on Artificial
  Intelligence}, volume~33, pages 7297--7304.

\bibitem[{Xiao et~al.(2021)Xiao, Pang, Lan, Wang, Shen, and
  Cheng}]{DBLP:conf/emnlp/XiaoPLWSC21}
Fei Xiao, Liang Pang, Yanyan Lan, Yan Wang, Huawei Shen, and Xueqi Cheng. 2021.
\newblock \href {https://aclanthology.org/2021.emnlp-main.195} {Transductive
  learning for unsupervised text style transfer}.
\newblock In \emph{Proceedings of the 2021 Conference on Empirical Methods in
  Natural Language Processing, {EMNLP} 2021, Virtual Event / Punta Cana,
  Dominican Republic, 7-11 November, 2021}, pages 2510--2521. Association for
  Computational Linguistics.

\bibitem[{Xu et~al.(2020)Xu, Crego, and Senellart}]{xu2020boosting}
Jitao Xu, Josep~M Crego, and Jean Senellart. 2020.
\newblock Boosting neural machine translation with similar translations.
\newblock In \emph{Proceedings of the 58th Annual Meeting of the Association
  for Computational Linguistics}, pages 1580--1590.

\bibitem[{Yang et~al.(2019)Yang, Hu, Qiu, Qu, Gao, Croft, Liu, Shen, and
  Liu}]{yang2019hybrid}
Liu Yang, Junjie Hu, Minghui Qiu, Chen Qu, Jianfeng Gao, W~Bruce Croft,
  Xiaodong Liu, Yelong Shen, and Jingjing Liu. 2019.
\newblock A hybrid retrieval-generation neural conversation model.
\newblock In \emph{Proceedings of the 28th ACM international conference on
  information and knowledge management}, pages 1341--1350.

\bibitem[{Zhang et~al.(2018)Zhang, Utiyama, Sumita, Neubig, and
  Nakamura}]{zhang2018guiding}
Jingyi Zhang, Masao Utiyama, Eiichiro Sumita, Graham Neubig, and Satoshi
  Nakamura. 2018.
\newblock Guiding neural machine translation with retrieved translation pieces.
\newblock In \emph{Proceedings of the 2018 Conference of the North American
  Chapter of the Association for Computational Linguistics: Human Language
  Technologies, Volume 1 (Long Papers)}, pages 1325--1335.

\bibitem[{Zhang et~al.(2021)Zhang, Sun, Gao, Fang, Brockett, Galley, Gao, and
  Dolan}]{zhang2021joint}
Yizhe Zhang, Siqi Sun, Xiang Gao, Yuwei Fang, Chris Brockett, Michel Galley,
  Jianfeng Gao, and Bill Dolan. 2021.
\newblock Joint retrieval and generation training for grounded text generation.
\newblock \emph{arXiv preprint arXiv:2105.06597}.

\bibitem[{Zhang et~al.(2020)Zhang, Chen, Wang, Utiyama, Sumita, Li, and
  Zhao}]{DBLP:conf/iclr/0001C0USLZ20}
Zhuosheng Zhang, Kehai Chen, Rui Wang, Masao Utiyama, Eiichiro Sumita, Zuchao
  Li, and Hai Zhao. 2020.
\newblock \href {https://openreview.net/forum?id=Byl8hhNYPS} {Neural machine
  translation with universal visual representation}.
\newblock In \emph{8th International Conference on Learning Representations,
  {ICLR} 2020, Addis Ababa, Ethiopia, April 26-30, 2020}. OpenReview.net.

\bibitem[{Zhechev and Van~Genabith(2010)}]{zhechev2010seeding}
Ventsislav Zhechev and Josef Van~Genabith. 2010.
\newblock Seeding statistical machine translation with translation memory
  output through tree-based structural alignment.
\newblock In \emph{Proceedings of the 4th Workshop on Syntax and Structure in
  Statistical Translation}, pages 43--51.

\bibitem[{Zheng et~al.(2021{\natexlab{a}})Zheng, Zhang, Guo, Huang, Chen, Luo,
  and Chen}]{zheng2021adaptive}
Xin Zheng, Zhirui Zhang, Junliang Guo, Shujian Huang, Boxing Chen, Weihua Luo,
  and Jiajun Chen. 2021{\natexlab{a}}.
\newblock Adaptive nearest neighbor machine translation.
\newblock \emph{arXiv preprint arXiv:2105.13022}.

\bibitem[{Zheng et~al.(2021{\natexlab{b}})Zheng, Zhang, Huang, Chen, Xie, Luo,
  and Chen}]{zheng2021non}
Xin Zheng, Zhirui Zhang, Shujian Huang, Boxing Chen, Jun Xie, Weihua Luo, and
  Jiajun Chen. 2021{\natexlab{b}}.
\newblock Non-parametric unsupervised domain adaptation for neural machine
  translation.
\newblock In \emph{Findings of the Association for Computational Linguistics:
  EMNLP 2021}, pages 4234--4241.

\bibitem[{Zhou et~al.(2018)Zhou, Prabhumoye, and Black}]{zhou2018dataset}
Kangyan Zhou, Shrimai Prabhumoye, and Alan~W Black. 2018.
\newblock A dataset for document grounded conversations.
\newblock \emph{arXiv preprint arXiv:1809.07358}.

\end{thebibliography}
\bibliographystyle{acl_natbib}

\end{document}